\documentclass[10pt,twocolumn,letterpaper]{article}

\usepackage{iccv}
\usepackage{times}
\usepackage{epsfig}
\usepackage{graphicx}
\usepackage{amsmath}
\usepackage{amssymb}
\usepackage{subcaption} 
\usepackage{booktabs} 
\usepackage[normalem]{ulem}  

\usepackage{arydshln} 
\usepackage{multirow} 
\usepackage{pifont}
\usepackage{comment}
\usepackage{authblk}

\usepackage[dvipsnames]{xcolor}
\definecolor{myblue}{rgb}{0.239,0.553,0.565}
\definecolor{myred}{rgb}{0.7,0.3,0.3}
\definecolor{mygreen}{rgb}{0.196,0.803,0.196}
\definecolor{myblack}{rgb}{0,0,0}

\newcommand{\green}[1]{\textcolor{mygreen}{#1}}

\newcommand{\myparagraph}[1]{\vspace{0pt}\paragraph{#1}}

\definecolor{mypink1}{rgb}{0.858, 0.188, 0.478}
\definecolor{mypink2}{RGB}{219, 48, 122}
\definecolor{mypink3}{cmyk}{0, 0.7808, 0.4429, 0.1412}
\definecolor{mygray}{gray}{0.6}

\definecolor{mygreen}{RGB}{0, 192, 0}
\definecolor{myred}{RGB}{255, 0, 0}

\newcommand{\cmark}{\textcolor{mygreen}{\ding{51}}}
\newcommand{\xmark}{\textcolor{myred}{\ding{55}}}

\makeatletter
\renewcommand{\paragraph}{%
  \@startsection{paragraph}{4}%
  {\z@}{1.0ex \@plus 1ex \@minus .2ex}{-1em}%
  {\normalfont\normalsize\bfseries}%
}
\makeatother
\usepackage[pagebackref=true,breaklinks=true,letterpaper=true,colorlinks,bookmarks=false]{hyperref}

\iccvfinalcopy 

\def\iccvPaperID{6} 

\ificcvfinal\fi

\begin{document}

\title{Online Detection of AI-Generated Images}

\author{David C. Epstein\hspace{5mm} Ishan Jain\hspace{5mm} Oliver Wang\hspace{5mm} Richard Zhang \\
Adobe Inc.}

\makeatletter
\let\@oldmaketitle\@maketitle
\renewcommand{\@maketitle}{\@oldmaketitle
\centering
\includegraphics[trim={0 .1cm 0 0}, width=1\linewidth]
{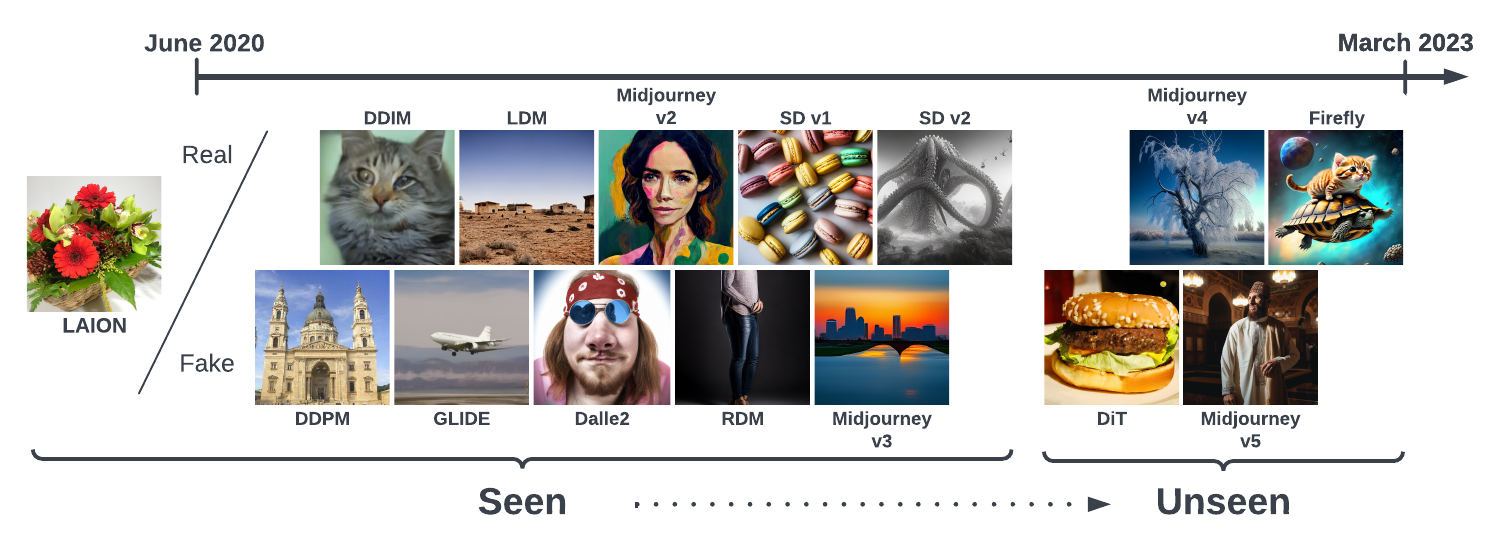}
\captionof{figure}{New AI-generated image models are released regularly. We train classifiers that distinguish real images from generated ones in an \emph{online} setting, whereby we add models to training in a simulated release order. We evaluate on \emph{unseen} model releases to see how well detectors might generalize into the future.}
\label{fig:teaser}
\bigskip}
\makeatother

\maketitle
\ificcvfinal\thispagestyle{empty}\fi

\begin{abstract}

With advancements in AI-generated images coming on a continuous basis, it is increasingly difficult to distinguish traditionally-sourced images (e.g., photos, artwork) from AI-generated ones. Previous detection methods study the generalization from a single generator to another in isolation. However, in reality, new generators are released on a streaming basis.
We study generalization in this setting, training on $N$ models and testing on the next $(N+k)$, following the historical release dates of well-known generation methods.
Furthermore, images increasingly consist of both real and generated components, for example through image inpainting.
Thus, we extend this approach to pixel prediction, demonstrating strong performance using automatically-generated inpainted data. In addition, for settings where commercial models are not publicly available for automatic data generation, we evaluate if pixel detectors can be trained solely on whole synthetic images.

\end{abstract}
\section{Introduction}

Since the advent of photography, image manipulation has been a source of misinformation and propaganda. In the digital age, tools to modify and distribute images have grown in effectiveness and scope. The emergence of AI-generated images has amplified these risks, democratizing the capacity to create realistic synthetic images indistinguishable from authentic images by human observers. Developing mechanisms to detect AI-generated images is critical to help address these concerns. 

Recent advances in image generation have increased the quality of synthetic outputs~\cite{ramesh2022hierarchical,saharia2022photorealistic,yu2022scaling,rombach2022high}, with new models and model variants released on a regular basis. This fast pace of development highlights \textit{future generalization} as a necessary component for an effective generative image detector.
Previous methods~\cite{wang2020cnn,chai2020makes} studying synthetic image detection and generalization suggest that different methods contain related, identifiable artifacts, showing that training a detector on \textit{one} generator generalizes to \textit{another}.

The true setting that we live in is online, with new methods coming out at a regular cadence.
While recent methods are driven by diffusion~\cite{sohl2015deep,ho2020denoising,song2020denoising}, many contain a fusion of elements, such as autoregressive components~\cite{yu2022scaling}, GAN-based upsampling or decoding~\cite{rombach2022high,kang2023scaling,goodfellow2014generative}, or perceptual losses~\cite{rombach2022high,zhang2018unreasonable}.
In addition, the amount of publicly available information for methods varies by model, from full open-source models with code and weights available to closed-source industry-based models, with only images available. Can the \textit{sum total} of available information on today's generators be used to detect tomorrow's models?

To answer this, we study detection and generalization emulating an \textit{online} setting.
We collect a dataset of 14 well-known generative models and simulate a real-world learning setting by training incremental CNN detectors, preserving the historical order by model release date.
We evaluate detector performance on images from both previously seen and unseen generative models, giving insight into the real-world performance of a live detector system.

In line with previous works, we find that current generative models continue to have exploitable cues that can be reliably detected when seen in the training set. A single classifier can even detect multiple generators simultaneously. Encouragingly, performance on \textit{unseen} models also increases, as the seen model history grows. This suggests that online training methodologies can exploit a collection of historical synthetic images to generalize to future unseen generative AI models.

In addition, as generative models start serving as the backbone of image \textit{editing} pipelines, images are increasingly a mixture of captured (real) and generated pixels.
One common method for generating such composites is by using generative models to ``inpaint'', or generate the pixels within the confines of a mask, replacing the original pixels behind it.
We extend our approach from the detection of wholly generated images to pixel-level predictions to detect such images and study generalization across Stable Diffusion~\cite{stablediffusion,stablediffusion2} versions and Adobe Firefly \cite{firefly}. 

We first investigate if synthetically generated inpainted images can be used to train a pixel-level detector. 
Next, we recognize that a challenge in pixel-level detection is the variation in available training data (e.g., ground truth masks), especially with closed models where we cannot synthetically generate inpainted images. 
For such situations, we show that one can leverage whole images to create pixel-wise predictions by using CutMix augmentation~\cite{yun2019cutmix}.

In summary our contributions are:\vspace{-2mm}
\begin{itemize}
\setlength\itemsep{.0em}
    \item We study detection of AI-generated images in an \textit{online} setting, utilizing 570,221 images from 14 generative methods.
    \item We study pixel-level prediction for inpainting applications, showing that CutMix augmentation improves performance in the absence of pixel-wise training data.
\end{itemize}
\section{Related work}

\begin{table*}
\begin{center}
\resizebox{1\linewidth}{!}{  
\begin{tabular}{llcccccrrr}
\toprule  
\multirow{2}{*}{\shortstack[l]{Generation\\ architecture}} & \multirow{2}{*}{Method/Dataset} & \multirow{2}{*}{\shortstack[c]{Training \\ set}} & \multicolumn{2}{c}{Method availability} & \multicolumn{2}{c}{Ordering} & \multicolumn{3}{c}{Dataset size} \\ \cmidrule(lr){4-5} \cmidrule(lr){6-7} \cmidrule(lr){8-10}
& & & Paper & Open-src & $\#$ & Date & Train & Val & Test \\
\hline 
Real images & LAION-400M~\cite{schuhmann2021laion} & -- & -- & -- & -- & -- & 179,900 & 22,479 & 22,490 \\ \cdashline{1-10}

\multirow{4}{*}{\shortstack[l]{Diffusion \\ U-net}} & Denoising Diffusion Prob. Model (DDPM)~\cite{ho2020denoising} & LSUN~\cite{yu2015lsun} & \cmark & \cmark & 1 & Jun 20 & 6,271 & 784 & 785 \\ 
& Denoising Diffusion Implicit Model (DDIM)~\cite{song2020denoising} & LSUN~\cite{yu2015lsun} & \cmark & \cmark & 2 & May 21 & 8,000 & 1,000 & 1,000 \\
& GLIDE~\cite{nichol2021glide} & Private & \cmark & \cmark & 3 & Dec 21 & 7,442 & 929 & 931 \\
& DALL$\cdot$E 2~\cite{ramesh2022hierarchical} & Private & \cmark & \xmark & 5 & Apr 22 & 2,000 & 954 & 2,000\\ \cdashline{1-10}

\multirow{4}{*}{\shortstack[l]{Diffusion \\ +Decoder}} & Latent Diffusion (LDM)~\cite{rombach2022high}  & LAION-400M~\cite{schuhmann2021laion} & \cmark & \cmark & 4 & Dec 21 & 8,172 & 1,021 & 1,022 \\
& Retrieval-Augmented Diffusion (RDM)~\cite{blattmann2022retrieval} & LAION-400M~\cite{schuhmann2021laion} & \cmark & \cmark & 7 & Jul 22 & 8,528 & 1,066 & 1,066 \\
& Stable Diffusion v1.1-v1.4~\cite{rombach2022high,stablediffusion} & LAION-2B~\cite{schuhmann2022laion} & \cmark & \cmark & 8 & Aug 22 & 34,508 & 3,807 & 3,838 \\
& Stable Diffusion 2.0(-v), 2.1(-v)~\cite{rombach2022high,stablediffusion2} & LAION-5B~\cite{schuhmann2022laion} & \cmark & \cmark & 10 & Nov 22 & 35,997 & 4,000 & 4,000 \\ \cdashline{1-10}
Diffusion U-Vit & Diffusion w/ Transformers (DiT)~\cite{peebles2022scalable} & ImageNet~\cite{russakovsky2015imagenet} & \cmark & \cmark & 11 & Dec 22 & 3,199 & 400 & 401 \\ \cdashline{1-10}
\multirow{5}{*}{\shortstack[l]{Unknown \\ (product release)}} & Midjourney v2~\cite{midjourney} & Unknown & \xmark & \xmark & 6 & Jul 22 & 42,875 & 5,358 & 5,359 \\ 
& Midjourney v3~\cite{midjourney} & Unknown & \cmark & \cmark & 9 & Nov 22 & 70,035 & 8,754 & 8,755 \\ 
& Midjourney v4~\cite{midjourney} & Unknown & \xmark & \xmark & 12 & Feb 23 & 100,000 & 10,000 & 76,122 \\
& Midjourney v5~\cite{midjourney} & Unknown & \xmark & \xmark & 13 & Mar 23 & 63,310 & 7,914 & 7,918 \\
& Adobe Firefly~\cite{firefly} & Unknown & \xmark & \xmark & 14 & Mar 23 & 15,525 & 2,070 & 3,105 \\
\bottomrule 
\vspace{-8mm}
\end{tabular}
}
\end{center}
\caption{\textbf{Online dataset.} We gather recent generative models. We show the methods, grouped by their image generation method. For our training experiments, we order them chronologically, recreating an online training scenario from recent history. As many of the generated methods are general text-to-image generators, we use a subset of LAION~\cite{schuhmann2021laion} to represent real images. Note that for some methods (Dall$\cdot$E 2, Midjourney, Firefly), the models are not open-sourced and only accessible through web interfaces, with no official architectural details available for some (Midjourney, Firefly).
}
\label{online-genai-table}
\end{table*}

\myparagraph{Generative modeling.} Generative models aim to model the distribution of data, given a set of samples. Early attempts, utilizing deep networks, include restricted Boltzmann machines (RBMs)~\cite{smolensky1986information} and deep Boltzmann machines (DBMs)~\cite{salakhutdinov2009deep}. Recent approaches include variational autoencoders (VAEs)~\cite{kingma2013auto}, autoregressive models~\cite{van2016pixel,van2016conditional}, normalizing flows~\cite{dinh2016density}, generative adversarial networks (GANs)~\cite{goodfellow2014generative}, and diffusion models~\cite{sohl2015deep,ho2020denoising,song2020denoising}.
The ProGAN/StyleGAN family~\cite{karras2019style,karras2020analyzing} has demonstrate photorealistic results, mostly focused on single-class generation. Such methods prompted the exploration of forensics techniques to detect synthetic imagery.
Recently, diffusion models have shown breakthrough results on arbitrary text-to-image generation, with multiple methods, such as Stable Diffusion (based on LDMs)~\cite{rombach2022high}, DALL$\cdot$E 2~\cite{ramesh2022hierarchical}, Imagen~\cite{saharia2022photorealistic}, released in quick succession. With the capability to now seemingly generate ``everything and anything'', the ability to distinguish real from synthetic has become increasingly challenging. 

\myparagraph{Detecting generated images.}
Media forensics on synthetic images from traditional tools has a long history~\cite{farid2009image,verdoliva2020media}, for example leveraging signals such as resampling artifacts~\cite{popescu2005exposing}, JPEG quantization~\cite{agarwal2017photo}, and shadows~\cite{kee2013exposing}, and detecting operations such as image splicing~\cite{zhou2018learning,huh2018fighting} or Photoshop warps~\cite{wang2019detecting}.
As deep generative methods have democratized synthesis of arbitrary image content, recent work has explored the ability of deep discriminative methods to detect such content~\cite{wang2020cnn,chai2020makes}, primarily in the context of GAN-based techniques~\cite{karras2019style,karras2020analyzing,brock2018large}.

A key question is how well detectors generalize to methods not seen during training. Wang et al.~\cite{wang2020cnn} find that a simple classifier, trained on one GAN model, can generalize to others, especially when using aggressive augmentations. Chai et al.~\cite{chai2020makes} demonstrate that even small patches contain sufficient cues for detection. Other features, for example, focusing on frequency cues~\cite{frank2020leveraging,liu2022detecting}, using co-occurrence matrices~\cite{nataraj2019detecting}, or even pretrained CLIP~\cite{radford2021learning} features, and techniques such as augmentation with representation mixing~\cite{bui2022repmix} are also effective. A common failure case when generalizing to a new generator is that while average precision is high, accuracy is low, demonstrating effective separation between real and fake classes but poor calibration. Ojha et al.~\cite{ojha2023towards} demonstrate that a simple nearest neighbors classifier improve accuracy, though at the cost of inference time. We build on the general observation that even baseline classifiers can generalize across generators, by studying and characterizing their behavior in an online setting.

Do images from recent diffusion methods still contain detectable cues? Recent work~\cite{corvi2022detection,coccomini2023detecting,sha2023defake} show that although GAN-based detectors do not generalize to diffusion methods~\cite{corvi2022detection}, diffusion models are detectable and exhibit some generalizability to each other.
We take these studies further by training a detector on 14 methods in an online fashion, simulating their release dates, and releasing an accompanying dataset of 570k images.

While these works detect \textit{whole} images, \text{local} prediction also offers important use-cases. For example, forensics methods have targeted edits from traditional tools, such as Photoshop warping~\cite{wang2019detecting} and image splicing~\cite{zhou2018learning,huh2018fighting}. Chai et al.~\cite{chai2020makes} show that patch-based classifiers can generate heatmaps for regions that contain more detectable cues. In this work, we investigate if inpainted regions can be localized. We show that even without direct access to inpainted examples, by using CutMix augmentation~\cite{yun2019cutmix}, whole images can be leveraged to produce pixel predictions.
\section{Online detection of AI-generated images}

\subsection{Dataset}

To begin, we collect generated images from 14 models, released between June 2020 and March 2023, shown in Table \ref{online-genai-table}.
These models reflect the rapid pace of advancement in realistic synthetic image generation, including academic papers~\cite{ho2020denoising,song2020denoising,nichol2021glide,rombach2022high,peebles2022scalable} as well as company releases~\cite{ramesh2022hierarchical,midjourney,firefly,stablediffusion,stablediffusion2}.
As the architecture of the generative model plays a large role in what features might be detectable, we group approaches by model architecture.

This dataset is useful to evaluate the generalization of detectors to unseen generative models. We use these images and the corresponding release dates of the source models to simulate an online learning setting. For publicly available models, we use the model release time from the repository. For products that can only be queried with an API (Midjourney~\cite{midjourney} and Firefly~\cite{firefly}), as the release date and details exact deployed version is not known, we use the date we query the model. In total, our dataset is composed of 570,221 images (405,862 train, 48,057 val, 116,302 test).

\myparagraph{Diffusion with U-Nets} First, we collect pixel diffusion methods, starting with DDPM~\cite{ho2020denoising}, DDIM~\cite{song2020denoising} and GLIDE~\cite{nichol2021glide}. These methods are accompanied by papers and source code releases (a smaller model version, in the case of GLIDE).
All of them use a U-Net~\cite{ronneberger2015u} with a diffusion-based objective~\cite{sohl2015deep} as their architecture. DDPM and DDIM train unconditional models on smaller datasets, while GLIDE performs text-to-image generation.

Additionally, we include DALL$\cdot$E 2~\cite{ramesh2022hierarchical}, a high-profile model from OpenAI.
During data collection time, the model was only available through web interface, which makes collecting a large-scale dataset challenging.
Instead, we scrapped the DALL$\cdot$E 2 Reddit, keeping images of $1024\times 1024$ (to filter out extraneous content, such as memes). Because the web interface generates a watermark that would be easily identifiable by a classifier, we crop out the bottom 16 pixels of the image.

\myparagraph{Diffusion with other architectures.} The next largest family of methods are latent diffusion models (LDMs), first introduced by Rombach et al.~\cite{rombach2022high} and subsequently used with retrieval-based augmentation~\cite{blattmann2022retrieval}. 
These methods use a U-Net to perform diffusion in a latent domain, and then decode the latent signal with a decoder, trained as part of a variational autoencoder~\cite{kingma2013auto}, in combination with a GAN~\cite{goodfellow2014generative} and LPIPS perceptual loss~\cite{zhang2018unreasonable}. LDMs were subsequently popularized by the release of Stable Diffusion~\cite{stablediffusion} and Stable Diffusion 2~\cite{stablediffusion2}, a scaled-up version of Latent Diffusion trained on large scale web data, containing multiple subversions.
Additionally, several methods propose to change the diffusion U-Nets with transformers based on ViT~\cite{dosovitskiy2020image}, which have been shown to have advantages in discriminative tasks~\cite{peebles2022scalable,bao2023worth}.

\begin{figure*}
	\centering
        \begin{subfigure}{0.47\linewidth}
            \includegraphics[width=\linewidth]{
            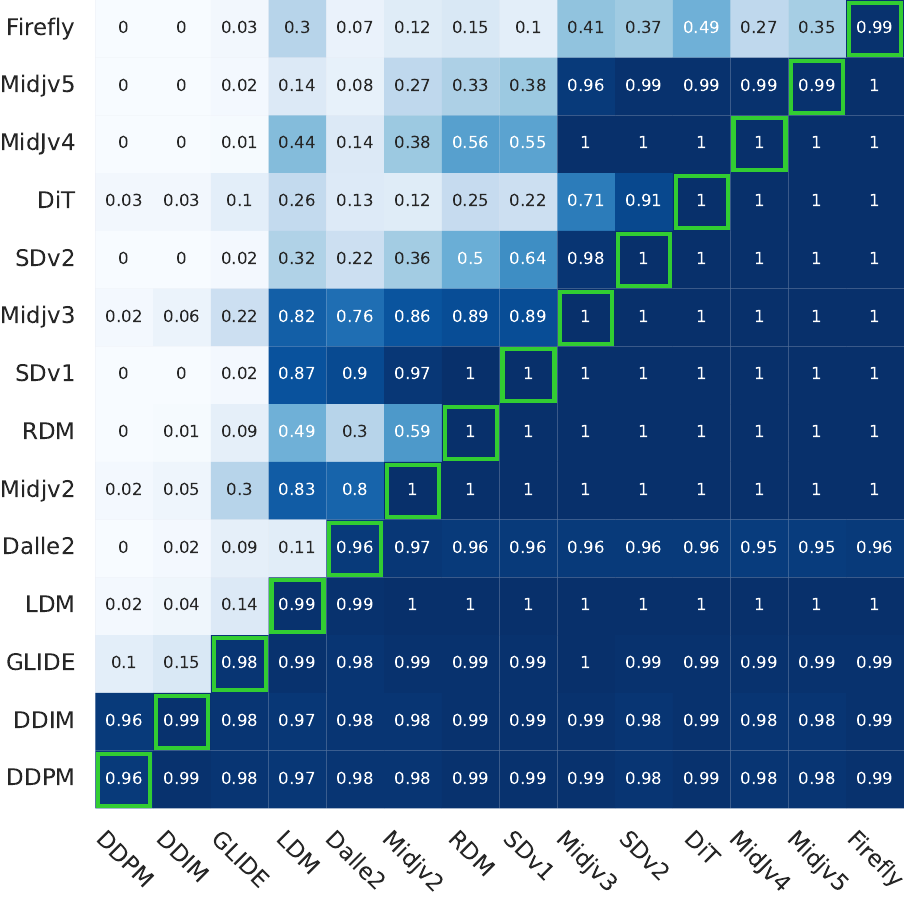
            }
		\caption{Accuracy}
		\label{fig:acc_a}
	\end{subfigure}
        \hspace{.5em} 
        \begin{subfigure}{0.47\linewidth}
            \includegraphics[width=\linewidth]{
            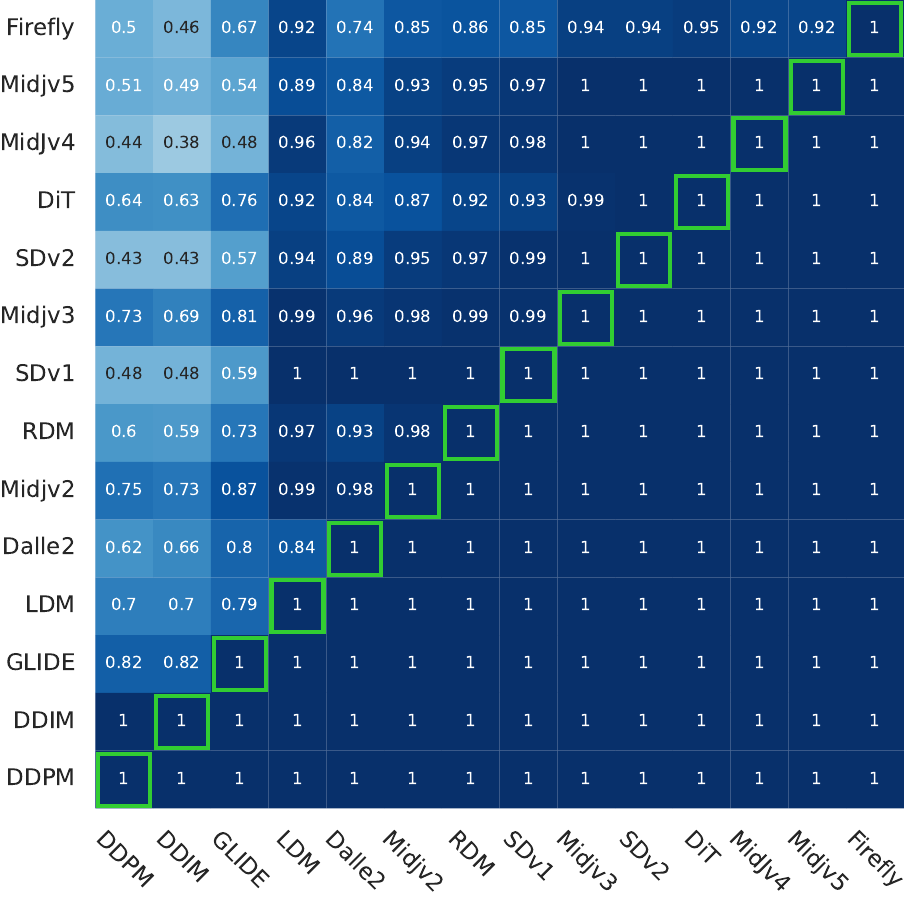
            }
		\caption{Area under Curve (AuC)}
		\label{fig:roc_b}
	\end{subfigure}
	\caption{\textbf{Online detector performance}. We order generative methods by their release date and train a real vs. fake classifier, in a progressive fashion (x-axis). We show (a) accuracy on synthetic images and (b) area under the curve (AuC) between real and fake images, across different test sets (y-axis). When testing on a generator that is in the training data, performance jumps to near 100$\%$ in both metrics, as seen on the \green{green} diagonal. After adding other generative sources, performance remains near 100$\%$ (bottom triangle). Even before being directly trained on (upper left triangle), accuracy and AuC on generative sources not yet seen by the detector increases, as the historical training set grows (for example, see SDv2 row). AuC considers both generative and non-generative images and is less sensitive to prediction thresholds, suggesting the online detector model rapidly learns features useful for differentiating AI generated images from non-generative images across a wide variety of models, although requiring some additional calibration (threshold selection) to improve accuracy.
 }
	\label{fig:online_matrices}
\end{figure*}

\myparagraph{Product releases.} With the success of image generation models, companies have released proprietary products. Such models make for interesting test cases, as one can speculate that they contain common elements from publicly available models, but such details are not publically disclosed. We sample images from Midjourney~\cite{midjourney} and Adobe Firefly~\cite{firefly}. We obtain Midjourney images by scraping the Discord API. As the model may be changing under the API and we do not know the true underlying model subversion, we date the models based on our scraping date.
We query Adobe Firefly images (without the watermarking used in their web interface).

\myparagraph{Prompt sourcing.} For GLIDE, LDM, RDM, and Firefly (all sets), and Stable Diffusion (train+val), we use prompts from DiffusionDB~\cite{wang2023diffusiondb}. For the Stable Diffusion test set, we use prompts from various web sources~\cite{openprompts,prompthero,midjourneyprompts,dalle2database}.We sample unique prompts, in order to not have overlap between train, validation, and test sets.

\subsection{Training details}

We progressively train a binary classifier with a cross-entropy loss to distinguish between naturally sourced ``real'' images and those generated by AI. We follow best practices from Wang et al.~\cite{wang2020cnn}, which show that a simple classifier can generalize across generators.
We use a common CNN architecture, ResNet-50~\cite{he2016deep}, pre-trained on ImageNet~\cite{russakovsky2015imagenet} as the backbone for the online detector model. 
The training sequence follows Table~\ref{online-genai-table} and Figure~\ref{fig:teaser}, simulating the real-world release dates of generative models. Each detector training step in the sequence continues from the previous model weights and incorporates all historical images seen to date. Release dates are determined by paper publication date, service launch announcement, or public release of model weights as relevant for the generative source.

Non-generated images sourced from LAION-400M are included at the start of training and remain a fixture throughout learning. 
During each training stage, we use a class-balanced random sampler to balance the distribution of generated and non-generated images over an epoch. Wang et al.~\cite{wang2019detecting} find that augmentations improve generalization. As such, we use a random 256$\times$256 crop randomly applying Gaussian blur (with probability $p=.01$), grayscale ($p=.05$), and invisible watermarks~\cite{invisible-watermark} ($p=.2$), a common feature of generative services. 
For evaluation, we center crop to 256$\times$256 with no augmentation.

\subsection{Results}

In Figure \ref{fig:online_matrices}, we show the results of our online detector.
The x-axis of the two heatmaps is the online model's training progress through time. Every column is a step in the learning progression, with a classifier trained on all models up to that cell, which are ordered by release date. For example, the ``LDM'' column includes images from DDPM, DDIM, GLIDE, and LDM.
The y-axis represents the test set, with model release date ordered from bottom to top. 
In this way, cells in the upper left triangle show \textit{generalization} performance of the online detector model on models, not yet included in the training.

\myparagraph{Metrics.} Figure \ref{fig:acc_a} shows \textbf{accuracy} on the synthetic images. Figure \ref{fig:roc_b} shows the \textbf{area under the curve (AuC)} between real and fake, sweeping over different thresholds, plotting the true positive and false positive rate, and taking the integral. The AuC metric shows how well the distributions are separated and is not sensitive to the classifier threshold. A high score indicates that the top most suspicious images can be reliably surfaced from a collection of images, useful in triaging scenarios. However, a common failure case of detectors on unseen distributions of synthetic images is classifying them as all real (low accuracy), as observed by Ojha et al~\cite{ojha2023towards}, even when AuC is high. Accuracy on real images typically remains near $100\%$. Thus, we show both AuC, which measures if the classifier has the right set of features, and synthetic accuracy, which measures if the classifier also has the right calibration.

\myparagraph{Does \textit{directly} training on a generator work?} Before investigating out-of-distribution generalization, we first perform a sanity check that a detector can be directly trained on a given generator. Observe the diagonal in Figure~\ref{fig:online_matrices}, outlined in green, which indicates the performance on a given generator directly after adding it to the training set.
Indeed, even a straightforward CNN-based classifier can differentiate from generated images and real images fairly well. Recall that while we are evaluating in-distribution, these are on a separate held-out set of images. AuC is at $100\%$ across all models, indicating perfect separation. Accuracy is near perfect, with worst case being $96\%$ accuracy. This test serves as a sanity check and follows conclusions from previous work~\cite{wang2020cnn}, that generators continue to have detectable cues, despite leaps in generative quality.

\myparagraph{Can a detector reliably detect \textit{multiple seen} generators simultaneously?} After adding in other datasets, performance on seen generators (bottom-right) remains at roughly the same accuracy level. The final column is trained on all our generators and produces near-perfect performance across the datasets. This indicates that a single detector can learn the different artifacts from generators, without running out of capacity.

\myparagraph{Can a detector generalize to \textit{unseen} generators?} Next, we investigate generalization ability and observe the upper-left of the matrix, testing on
not-yet-released models.
If classifiers do not generalize at all, we would see $0\%$ accuracy and chance 0.5 $AuC$ in the upper triangle. Previous work~\cite{wang2020cnn,chai2020makes} observe that classifiers do generalize, and indeed, we see values higher than chance, even before they are directly trained on, however performance degrades over time, indicating that online training will still be required as new models are released. 

Looking at the values one row above the diagonal indicates performance of training on $N$ models and testing on the $(N+1)^\text{th}$. Many of the methods -- DDIM, Midjourney (v2 to v5), RDM, Stable Diffusion (v1 and v2), DiT -- have AuC at $0.98$ or above, indicating that the classifier contains the right features early on in training, after DALL$\cdot$E 2 is added, the 5$^\text{th}$ model.

The accuracy scores are less consistent. For example, GLIDE achieves $15\%$ accuracy while having perfect AuC, indicating that additional calibration is needed.
The last model, Adobe Firefly, has relatively low accuracy, even when training on all previous models, indicating significant differences to previous models.

\myparagraph{Which generators have high impact on one another?}
It is also interesting to note that related models are correlated. 
For example, adding LDMs~\cite{rombach2022high} has outsized influence on a number of models, as indicated by the step up in performance in the AuC (Fig~\ref{fig:roc_b}). This includes direct descendants of RDMs~\cite{blattmann2022retrieval}, Stable Diffusion versions~\cite{stablediffusion,stablediffusion2}, as well as Midjourney, an unknown model.

Viewing the accuracy (Fig~\ref{fig:acc_a}), Midjourney v3 has large influence over subsequent versions of v4 and v5, indicating a large version change compared to Midjourney v2. As specific details of the model are not known, this sheds some light into the underlying model changes.

\begin{figure*}
    \includegraphics[width=\linewidth]{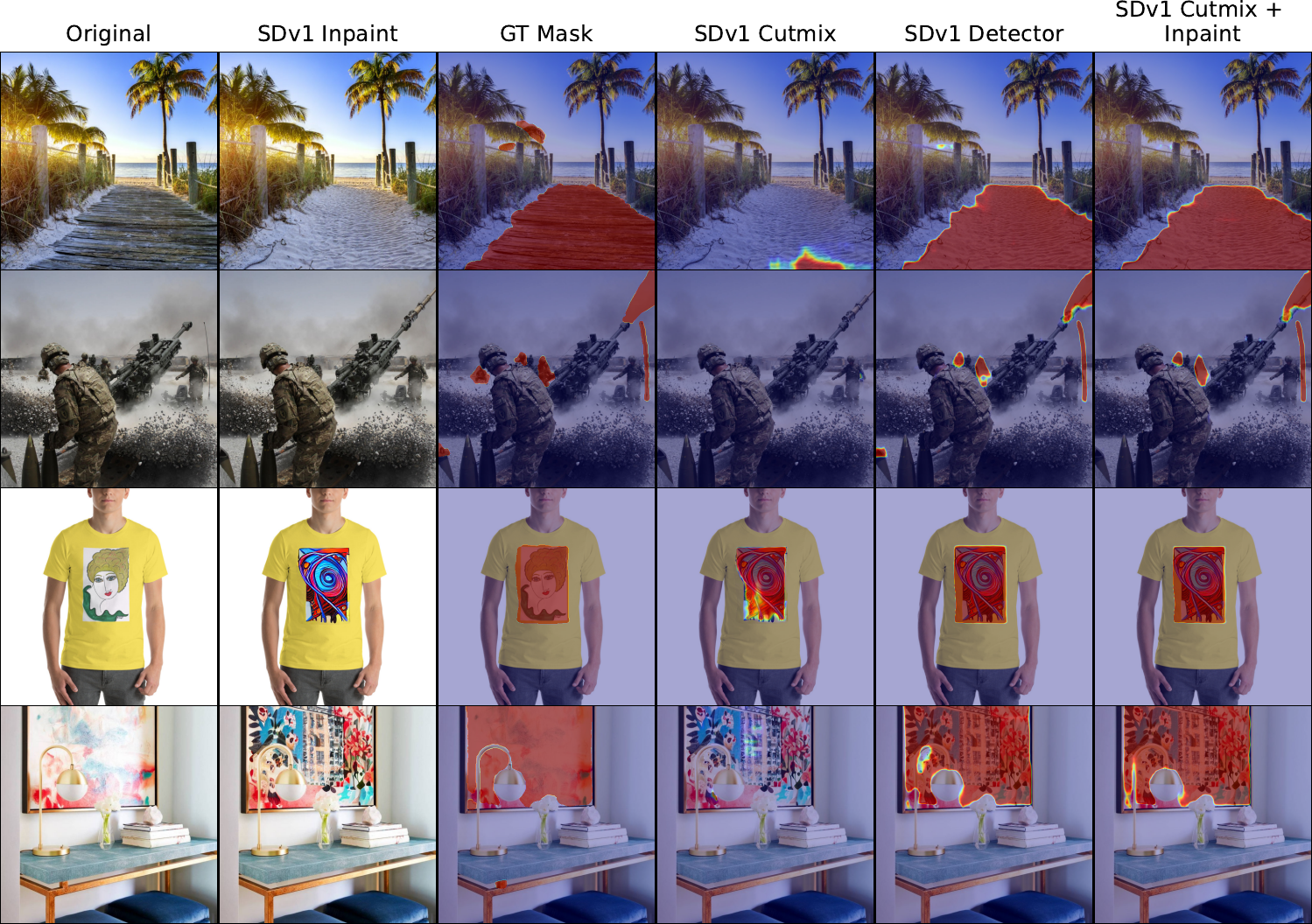}
\caption{\textbf{Inpainting detection and generalization (SDv1).}
Here we highlight the successful detection of SDv1 inpainted regions on LAION images. We also compare pixel detectors trained by various methods. While a SDv1 CutMix detector does provide some predictive power, inpainted example images are required in training for high accuracy.
}
 \label{fig:inpainting-sdv1}
\end{figure*}

\begin{figure*}
        \includegraphics[width=\linewidth]{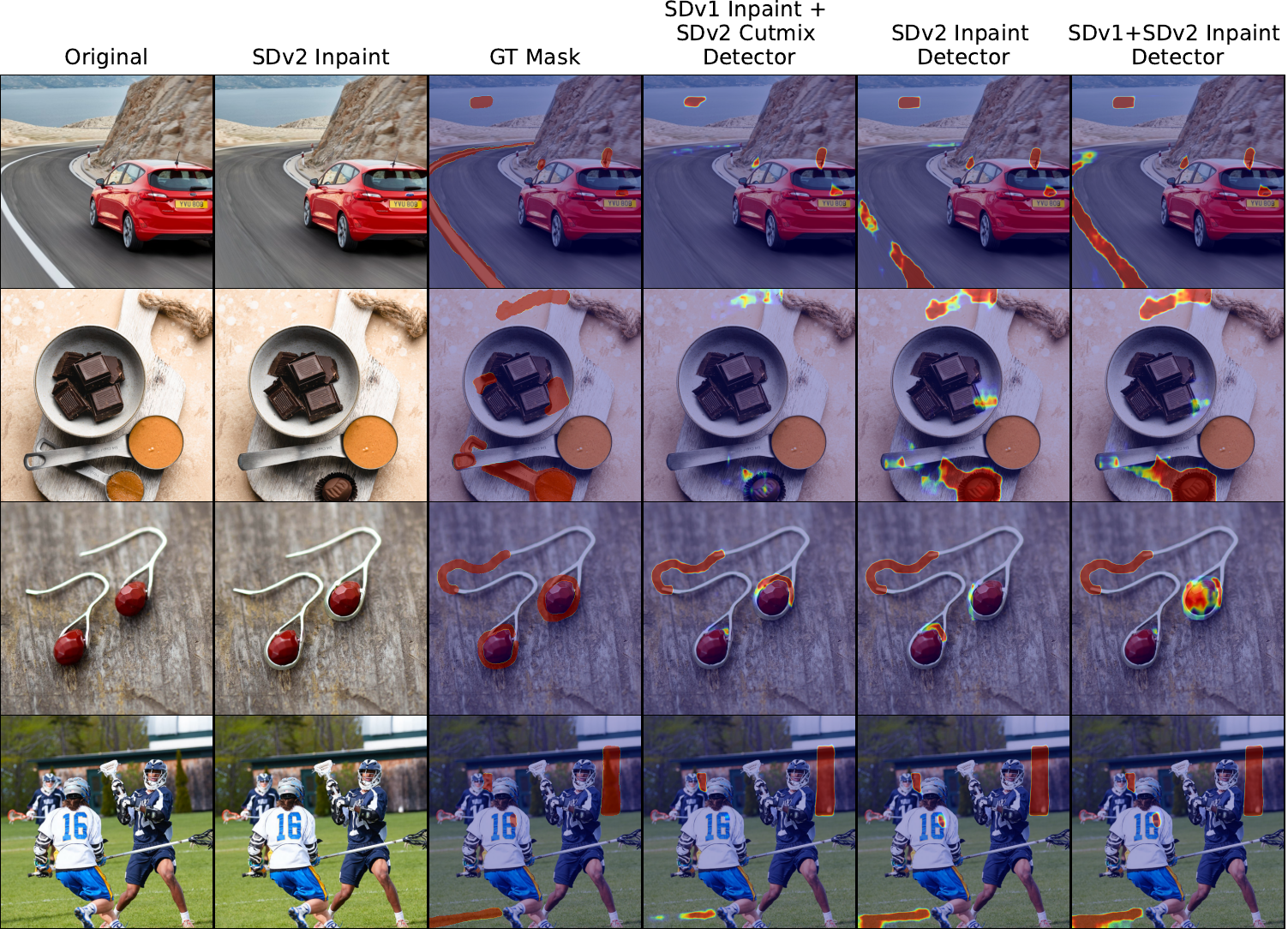}
\caption{\textbf{Inpainting detection and generalization (SDv2).} We can clearly see the progression in accuracy on SDv2 inpainted LAION images as higher quality data and an additional model source is introduced in pixel detector training. }
    \label{fig:inpainting-sdv2}
\end{figure*}

\begin{figure*}
 \begin{subfigure}{0.24\textwidth}
         \includegraphics[width=\textwidth]{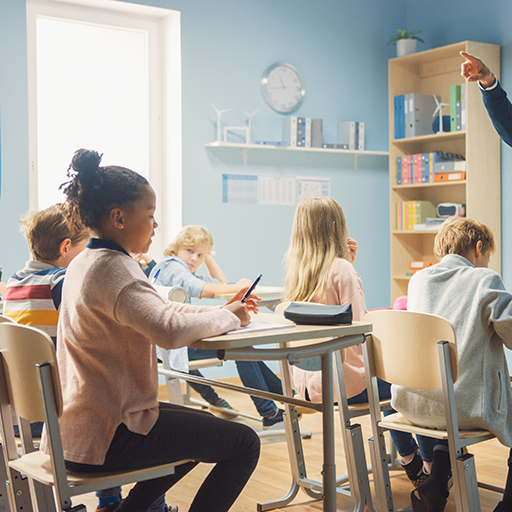}
     \caption{Original}
     \label{fig:a}
 \end{subfigure}
 \hfill
 \begin{subfigure}{0.24\textwidth}
     \includegraphics[width=\textwidth]{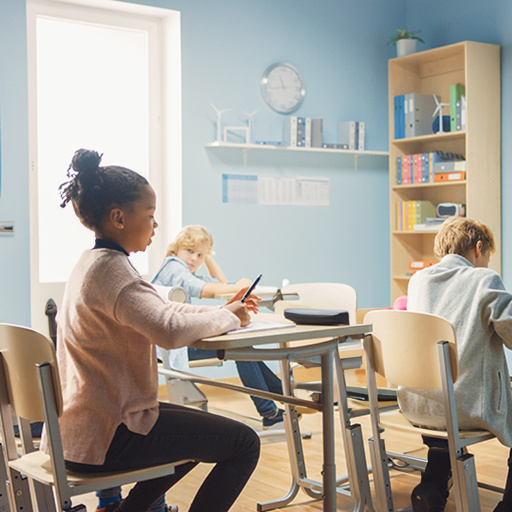}
     \caption{Firefly Inpainted Result}
     \label{fig:b}
 \end{subfigure}
\hfill
 \begin{subfigure}{0.24\textwidth}
     \includegraphics[width=\textwidth]{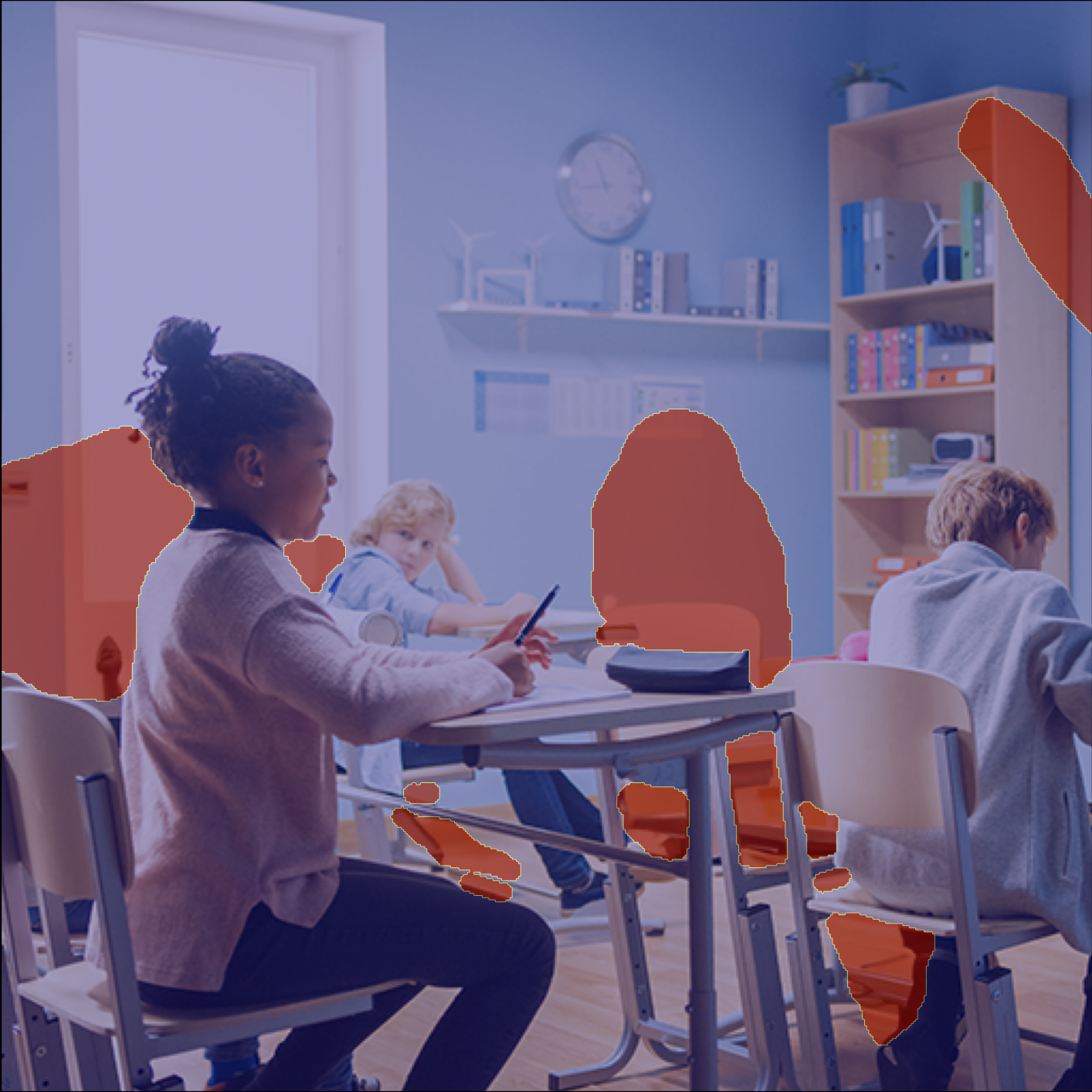}
     \caption{Ground Truth Mask}
     \label{fig:c}
 \end{subfigure}
 \hfill
 \begin{subfigure}{0.24\textwidth}
     \includegraphics[width=\textwidth]{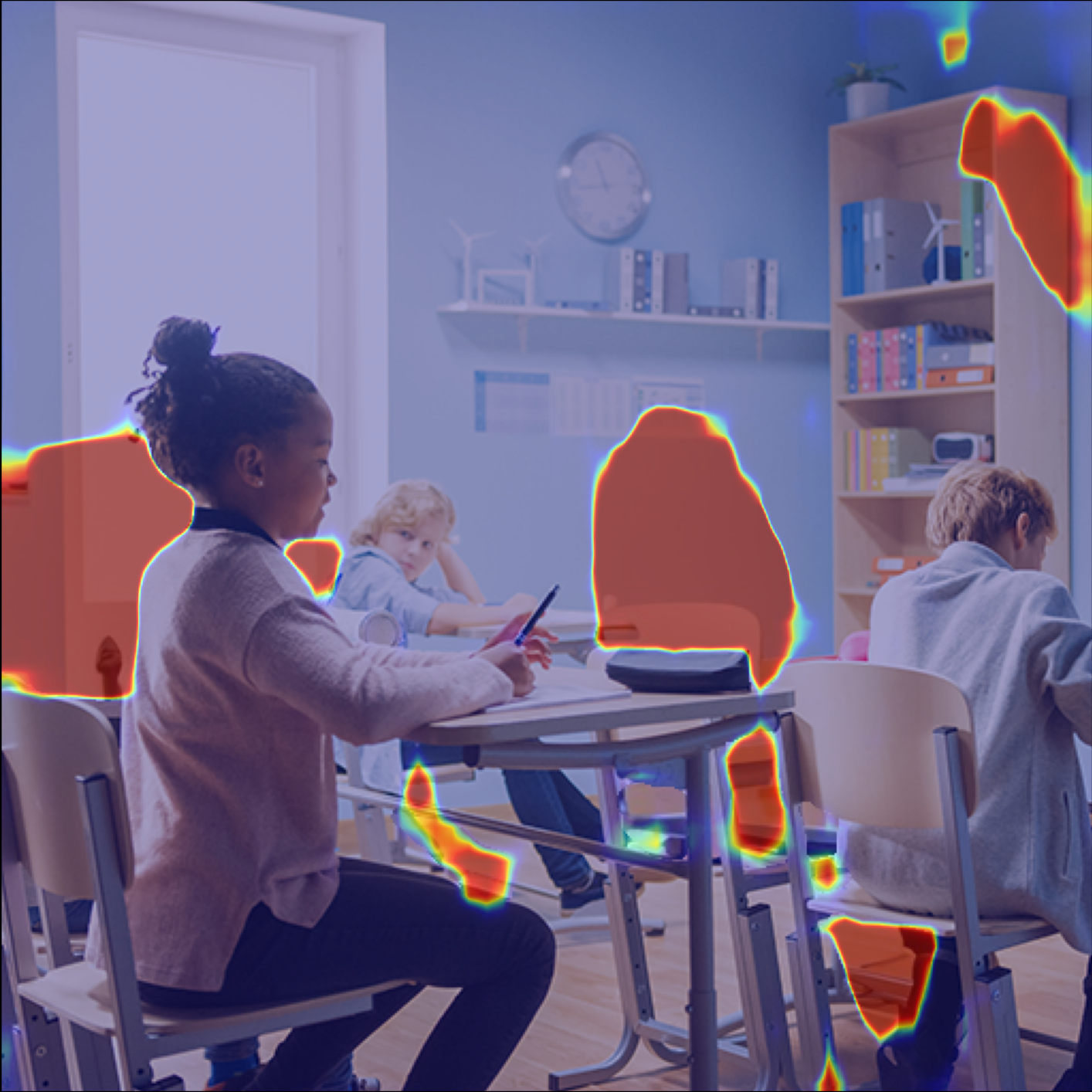}
     \caption{Detection}
     \label{fig:d}
 \end{subfigure}

 \caption{\textbf{Generative AI pixel detection test.} A detector model trained on inpainted images successfully captures the generated pixels from Adobe Firefly's inpainting model.}
 \label{fig:adobe_firefly_inpaint_detection}
\end{figure*}

\section{Detection of Generative Inpainting}

Outputs from generative models are increasingly used in editing pipelines, where final images are a composite of both AI-generated pixels and traditionally sourced images. 
One common way these images are generated is with ``inpainting'', where a masked region of an image is seamlessly filled with generated content. 
To evaluate how well we can detect these types of edits, we create a dataset from Adobe Firefly and Stable Diffusion's inpainting models.
As ground truth masks may be difficult to acquire in real-world settings, e.g., with closed models, we examine how one can leverage whole images to get pixel-wise labels, using CutMix~\cite{yun2019cutmix} augmentation.
We also investigate generalization by the inclusion of images from multiple inpainting models. 

\subsection{Dataset}

We create three inpainting datasets, using Stable Diffusion (v1 and v2)~\cite{stablediffusion,stablediffusion2} and Adobe Firefly~\cite{firefly}. 
Table \ref{inpaint-table} summarizes model release dates, input source, masked pixel distribution, and dataset size.
We sample input images and corresponding prompts from the LAION-400M Dataset\cite{schuhmann2021laion}.
We resize images to 512 pixels on the short side and center crop and generate a mask. We create masks covering 15 to 35$\%$ of each image, with random overlapping strokes and shapes.
In order to preserve the fidelity of the non-masked region and isolate the generated pixels from the original, we copy the original image back into the non-masked region. Importantly, we do not observe this approach to cause a visible seam.

For whole image sets, we use LAION images from Table \ref{inpaint-table} and sample the desired generative models (in equal size) from the online dataset previously described in Table \ref{online-genai-table}.

\begin{table}[t]
\begin{center}
\resizebox{1\linewidth}{!}{ 
\begin{tabular}{*7c} 
\toprule
Inpaint Model & Release Date & Input & Mask \% & \textit{Train} & \textit{Val} & \textit{Test} \\ 
\hline 
SDv1 & Oct 22 & LAION & 15-35 & 9375 & 1250 & 1875 \\
SDv2 & Nov 22 & LAION & 15-35 & 9375 & 1250 & 1875 \\
Firefly & Mar 23 & LAION & 15-35 & 9091 & 1206 & 1813 \\
\bottomrule
\vspace{-8mm}
\end{tabular}
}
\end{center}
\caption{\textbf{Inpainting dataset.} We investigate pixel-wise generative detection by creating a dataset of inpainting results, using Stable Diffusion and Adobe Firefly.}
\label{inpaint-table}
\end{table}

\subsection{Training details}

Similarly to whole image detection, we study generalization using a standard Fully Convolutional Network (FCN)~\cite{long2015fully} with a ResNet-50~\cite{he2016deep} architecture, utilizing weighted Dice loss~\cite{Sudre_2017} to address data imbalance.
At training, we test schemes with the assumption that an inpainting API is or is not available.
When not available, we test whether \textit{whole} images can be leveraged for pixel-wise labels. We test CutMix blending~\cite{yun2019cutmix}, a naive approach of simply cutting and pasting blocks of one image into another~\cite{yun2019cutmix}. When an inpainting API is available, we test with synthetically-generated inpainted images with random masks, described in the section above. We also test if adding whole images with CutMix can help in this situation.

\begin{table}
\begin{center}
\resizebox{1\linewidth}{!}{  
\begin{tabular}{llcccc}

\toprule  
\multicolumn{2}{l}{ \multirow{2}{*}{Training Data}} & \multicolumn{2}{c}{SDv1 Inpainting} & \multicolumn{2}{c}{Whole image} \\ \cmidrule(lr){3-4} \cmidrule(lr){5-6}

 &  & accuracy & f1 & accuracy & f1  \\

\hline 
SDv1 whole image &  & 0.7755 & 0.1807 & 0.9979 & 0.9968  \\
SDv1 cutmix &  & 0.8359 & 0.4882 & 0.9960 & 0.9941  \\
SDv1 inpaint  &  & 0.9902 & 0.9795 & 0.9918 & 0.9878  \\
SDv1 inpaint + cutmix  &  & \textbf{0.9920} & \textbf{0.9832} & \textbf{0.9996} & \textbf{0.9994}  \\
\bottomrule 
\end{tabular}
}
\end{center}
\vspace{-3mm}
\caption{\textbf{SDv1 inpainting detection.} Compared to using whole images only, we observe improvement when training with CutMix augmentation, and another jump when trained on inpainted images, with the best model using both. Additionally, we see these methods retain accuracy on whole image detection in a test set comprising both LAION and pure SDv1 images.}
\label{sdv1-inpaint-detection-table}
\end{table}

\begin{table}
\begin{center}
\resizebox{1\linewidth}{!}{ 
\begin{tabular}{lcccc} 
\toprule

Training Data &  accuracy & precision & recall & f1 \\ [0.5ex] 
\hline 
SDv1 inpaint & 0.9288 & 0.9776 & 0.7173 & 0.8275 \\
SDv1 inpaint + SDv2 cutmix & 0.9600 & 0.9781 & 0.8509 & 0.9101 \\
SDv2 inpaint & 0.9872 & 0.9766 & 0.9693 & 0.9730 \\

SDv1 inpaint + SDv2 inpaint & \textbf{0.9892} & \textbf{0.9811} & \textbf{0.9733} & \textbf{0.9772} \\
\bottomrule
\end{tabular}
}
\end{center}
\vspace{-3mm}
\caption{\textbf{SDv2 inpainting generalization.} A pixel detector trained on SDv1 inpainting images generalizes well to SDv2 inpainted images (1$^\text{st}$ row). When SDv2 whole images are added in training via CutMix, performance rises (2$^\text{nd}$ row), approaching the performance of a detector trained directly on SDv2 inpainted images (3$^\text{rd}$ row).}
\label{sdv2-inpaint-generalization-table}
\end{table}

\begin{table}
\begin{center}
\resizebox{1\linewidth}{!}{ 

\begin{tabular}{lcccc} 
\toprule

Training Data &  accuracy & precision & recall & f1 \\ [0.5ex] 
\hline 
Firefly cutmix  & 0.8250  & 0.9376 & 0.2870 & 0.4395 \\
SDv1 + SDv2 inpaint + Firefly cutmix & 0.9511 & 0.9749 & 0.8163 & 0.8886 \\
Firefly inpaint & 0.9811 & 0.9734 & 0.9469 &  0.9600 \\
SDv1 + SDv2 + Firefly inpaint  & \textbf{0.9891} & \textbf{0.9805} & \textbf{0.9740} & \textbf{0.9772} \\
\bottomrule
\end{tabular}
}
\end{center}
\vspace{-3mm}
\caption{\textbf{Firefly inpaint generalization.}  We see an increase in pixel accuracy when SDv1 and SDv2-inpainted images are added to Firefly CutMix detector training (1$^\text{st}$ to 2$^\text{nd}$ row), with performance relatively close to that achieved by directly training on Firefly inpainted images (3$^\text{rd}$ row).
}
\label{firefly-inpaint-generalization-table}
\end{table}

\subsection{Results}

\myparagraph{Can we create a pixel detector?} Table~\ref{sdv1-inpaint-detection-table} shows results on training and detecting SDv1. When only using whole images, we show that using CutMix improves performance ($77.6\% \rightarrow 83.6\%$ accuracy). In Figure~\ref{fig:inpainting-sdv1}, we see that CutMix can catch some inpainted regions, though far from perfect. When an inpainting API is available, we show that synthetically generated samples can greatly improve accuracy ($99.0\%$), with small improvements adding CutMix ($99.2\%$).
In Figure~\ref{fig:adobe_firefly_inpaint_detection}, we show a qualitative example a Firefly inpainting detector, trained on synthetically-generated Firefly samples.

\myparagraph{Can a pixel detector be used in conjunction with a whole image detector?}
Note that performance on detecting whole images is nearly perfect $>$99.0$\%$ across all cases, indicating the model is not simply relying on boundary artifacts. 

\myparagraph{Does a pixel detector generalize across models?}
In Table~\ref{sdv2-inpaint-generalization-table}, we measure performance on SDv2-generated inpainted images on detector methods, and if SDv1 can be leveraged to improve performance. We follow a progression from highly limited training data to most abundant.
We see in the most limited case, where the detector model only has access to images from the previous version of SDv1, a high-performance accuracy of 92.9\%, showing generalization at least in the case of closely related model sources. 
Leveraging SDv2, even with just whole images using CutMix, improves performance (96\%). 
This is approaching the performance of a stand-alone detector model trained directly on SDv2 inpainted images (98.7\%). Using the previous version of SDv1 further improves performance (98.9$\%$). In Figure~\ref{fig:inpainting-sdv2}, we show qualitative examples of this progressive improvement.

While SDv1$\rightarrow$SDv2 studies generalization across closely related models, in Table \ref{firefly-inpaint-generalization-table}, we study generalization across unrelated models, (SDv1+SDv2) $\rightarrow$Firefly.
We see that leveraging inpainted SDv1+SDv2 images improves performance, both compared to training only with Firefly whole images using CutMix (82.5$\rightarrow$95.1) and with inpainted Firefly images available (98.1$\rightarrow$98.9).
Our results indicate that generated images contain sufficient cues even at a local level that can be detected. Furthermore, incorporating previously seen models can improve accuracy.

\section{Conclusions and Limitations}
We conduct experiments to investigate how well classifiers can detect AI-generated images in a simulated online framework. 
We see that classifiers do generalize to unseen models, although when there are major architectural changes, performance drops substantially.
These experiments suggest that a vigilant classifier, regularly retrained on new generators, has the opportunity to detect future, unreleased models, as long as they are architecturally similar. 

While our dataset highlights some key advancements in generative methods, it does not comprise of all models in this time period.
Methods such as Imagen~\cite{saharia2022photorealistic}, Parti~\cite{yu2022scaling}, Muse~\cite{chang2023muse}, and GigaGAN~\cite{kang2023scaling} offer high-quality generations with alternative architectures but do not have public APIs or source code released.
Exploring detection for such methods is an interesting direction if the models become available.
In addition, we study online generalization with simple architectures (ResNet~\cite{he2016deep} and FCN~\cite{long2015fully}), following best training practices~\cite{wang2020cnn}. Further improving generalization in an online setting is an important future area of work.

\myparagraph{Acknowledgements}
\noindent We thank Sheng-Yu Wang and Alexei A. Efros for helpful discussion, Charlie Scheinost and Josh Arteaga for their support and discussion, and Deepti Clark for help gathering inpainting data.



{\small
\bibliographystyle{ieee_fullname}
\bibliography{egbib}

\begin{thebibliography}{10}\itemsep=-1pt

\bibitem{firefly}
Adobe firefly.
\newblock \url{https://firefly.adobe.com/}.

\bibitem{dalle2database}
Dall·e 2 image database.
\newblock \url{https://dalle2.gallery}.

\bibitem{midjourney}
Midjourney.
\newblock \url{https://www.midjourney.com/}.

\bibitem{prompthero}
Prompt hero.
\newblock \url{http://prompthero.com/}.

\bibitem{agarwal2017photo}
Shruti Agarwal and Hany Farid.
\newblock Photo forensics from jpeg dimples.
\newblock In {\em 2017 IEEE workshop on information forensics and security
  (WIFS)}, pages 1--6. IEEE, 2017.

\bibitem{openprompts}
Krea AI.
\newblock Open prompts.
\newblock \url{https://github.com/krea-ai/open-prompts}.

\bibitem{stablediffusion2}
Stability AI.
\newblock Stable diffusion version 2.
\newblock \url{https://github.com/Stability-AI/stablediffusion}.

\bibitem{bao2023worth}
Fan Bao, Shen Nie, Kaiwen Xue, Yue Cao, Chongxuan Li, Hang Su, and Jun Zhu.
\newblock All are worth words: A vit backbone for diffusion models.
\newblock In {\em CVPR}, 2023.

\bibitem{blattmann2022retrieval}
Andreas Blattmann, Robin Rombach, Kaan Oktay, Jonas M{\"u}ller, and Bj{\"o}rn
  Ommer.
\newblock Retrieval-augmented diffusion models.
\newblock {\em NeurIPS}, 2022.

\bibitem{brock2018large}
Andrew Brock, Jeff Donahue, and Karen Simonyan.
\newblock Large scale gan training for high fidelity natural image synthesis.
\newblock In {\em ICLR}, 2019.

\bibitem{bui2022repmix}
Tu Bui, Ning Yu, and John Collomosse.
\newblock Repmix: Representation mixing for robust attribution of synthesized
  images.
\newblock In {\em ECCV}, 2022.

\bibitem{chai2020makes}
Lucy Chai, David Bau, Ser-Nam Lim, and Phillip Isola.
\newblock What makes fake images detectable? understanding properties that
  generalize.
\newblock In {\em ECCV}, 2020.

\bibitem{chang2023muse}
Huiwen Chang, Han Zhang, Jarred Barber, AJ Maschinot, Jose Lezama, Lu Jiang,
  Ming-Hsuan Yang, Kevin Murphy, William~T Freeman, Michael Rubinstein, et~al.
\newblock Muse: Text-to-image generation via masked generative transformers.
\newblock In {\em PMLR}, 2023.

\bibitem{coccomini2023detecting}
Davide~Alessandro Coccomini, Andrea Esuli, Fabrizio Falchi, Claudio Gennaro,
  and Giuseppe Amato.
\newblock Detecting images generated by diffusers.
\newblock In {\em arXiv}, 2023.

\bibitem{corvi2022detection}
Riccardo Corvi, Davide Cozzolino, Giada Zingarini, Giovanni Poggi, Koki Nagano,
  and Luisa Verdoliva.
\newblock On the detection of synthetic images generated by diffusion models.
\newblock In {\em IEEE International Conference on Acoustics, Speech and Signal
  Processing (ICASSP)}, 2023.

\bibitem{dinh2016density}
Laurent Dinh, Jascha Sohl-Dickstein, and Samy Bengio.
\newblock Density estimation using real nvp.
\newblock In {\em ICLR}, 2017.

\bibitem{dosovitskiy2020image}
Alexey Dosovitskiy, Lucas Beyer, Alexander Kolesnikov, Dirk Weissenborn,
  Xiaohua Zhai, Thomas Unterthiner, Mostafa Dehghani, Matthias Minderer, Georg
  Heigold, Sylvain Gelly, et~al.
\newblock An image is worth 16x16 words: Transformers for image recognition at
  scale.
\newblock In {\em ICLR}, 2021.

\bibitem{farid2009image}
Hany Farid.
\newblock Image forgery detection.
\newblock {\em IEEE Signal processing magazine}, 26(2):16--25, 2009.

\bibitem{frank2020leveraging}
Joel Frank, Thorsten Eisenhofer, Lea Sch{\"o}nherr, Asja Fischer, Dorothea
  Kolossa, and Thorsten Holz.
\newblock Leveraging frequency analysis for deep fake image recognition.
\newblock In {\em ICML}, 2020.

\bibitem{goodfellow2014generative}
Ian Goodfellow, Jean Pouget-Abadie, Mehdi Mirza, Bing Xu, David Warde-Farley,
  Sherjil Ozair, Aaron Courville, and Yoshua Bengio.
\newblock Generative adversarial nets.
\newblock In {\em NIPS}, 2014.

\bibitem{he2016deep}
Kaiming He, Xiangyu Zhang, Shaoqing Ren, and Jian Sun.
\newblock Deep residual learning for image recognition.
\newblock In {\em CVPR}, 2016.

\bibitem{ho2020denoising}
Jonathan Ho, Ajay Jain, and Pieter Abbeel.
\newblock Denoising diffusion probabilistic models.
\newblock In {\em NeurIPS}, 2020.

\bibitem{huh2018fighting}
Minyoung Huh, Andrew Liu, Andrew Owens, and Alexei~A Efros.
\newblock Fighting fake news: Image splice detection via learned
  self-consistency.
\newblock In {\em ECCV}, 2018.

\bibitem{kang2023scaling}
Minguk Kang, Jun-Yan Zhu, Richard Zhang, Jaesik Park, Eli Shechtman, Sylvain
  Paris, and Taesung Park.
\newblock Scaling up gans for text-to-image synthesis.
\newblock In {\em CVPR}, 2023.

\bibitem{karras2019style}
Tero Karras, Samuli Laine, and Timo Aila.
\newblock A style-based generator architecture for generative adversarial
  networks.
\newblock In {\em CVPR}, 2019.

\bibitem{karras2020analyzing}
Tero Karras, Samuli Laine, Miika Aittala, Janne Hellsten, Jaakko Lehtinen, and
  Timo Aila.
\newblock Analyzing and improving the image quality of stylegan.
\newblock In {\em CVPR}, 2020.

\bibitem{kee2013exposing}
Eric Kee, James~F O'Brien, and Hany Farid.
\newblock Exposing photo manipulation with inconsistent shadows.
\newblock {\em ACM Transactions on Graphics (ToG)}, 32(3):1--12, 2013.

\bibitem{kingma2013auto}
Diederik~P Kingma and Max Welling.
\newblock Auto-encoding variational bayes.
\newblock In {\em ICLR}, 2014.

\bibitem{liu2022detecting}
Bo Liu, Fan Yang, Xiuli Bi, Bin Xiao, Weisheng Li, and Xinbo Gao.
\newblock Detecting generated images by real images.
\newblock In {\em ECCV}, 2022.

\bibitem{long2015fully}
Jonathan Long, Evan Shelhamer, and Trevor Darrell.
\newblock Fully convolutional networks for semantic segmentation.
\newblock In {\em CVPR}, 2015.

\bibitem{nataraj2019detecting}
Lakshmanan Nataraj, Tajuddin~Manhar Mohammed, Shivkumar Chandrasekaran, Arjuna
  Flenner, Jawadul~H Bappy, Amit~K Roy-Chowdhury, and BS Manjunath.
\newblock Detecting gan generated fake images using co-occurrence matrices.
\newblock {\em arXiv preprint arXiv:1903.06836}, 2019.

\bibitem{nichol2021glide}
Alex Nichol, Prafulla Dhariwal, Aditya Ramesh, Pranav Shyam, Pamela Mishkin,
  Bob McGrew, Ilya Sutskever, and Mark Chen.
\newblock Glide: Towards photorealistic image generation and editing with
  text-guided diffusion models.
\newblock In {\em ICML}, 2022.

\bibitem{ojha2023towards}
Utkarsh Ojha, Yuheng Li, and Yong~Jae Lee.
\newblock Towards universal fake image detectors that generalize across
  generative models.
\newblock In {\em CVPR}, 2023.

\bibitem{peebles2022scalable}
William Peebles and Saining Xie.
\newblock Scalable diffusion models with transformers.
\newblock In {\em ICCV}, 2022.

\bibitem{popescu2005exposing}
Alin~C Popescu and Hany Farid.
\newblock Exposing digital forgeries by detecting traces of resampling.
\newblock {\em IEEE Transactions on signal processing}, 53(2):758--767, 2005.

\bibitem{radford2021learning}
Alec Radford, Jong~Wook Kim, Chris Hallacy, Aditya Ramesh, Gabriel Goh,
  Sandhini Agarwal, Girish Sastry, Amanda Askell, Pamela Mishkin, Jack Clark,
  et~al.
\newblock Learning transferable visual models from natural language
  supervision.
\newblock In {\em ICML}, 2021.

\bibitem{ramesh2022hierarchical}
Aditya Ramesh, Prafulla Dhariwal, Alex Nichol, Casey Chu, and Mark Chen.
\newblock Hierarchical text-conditional image generation with clip latents.
\newblock {\em arXiv preprint arXiv:2204.06125}, 2022.

\bibitem{rombach2022high}
Robin Rombach, Andreas Blattmann, Dominik Lorenz, Patrick Esser, and Bj{\"o}rn
  Ommer.
\newblock High-resolution image synthesis with latent diffusion models.
\newblock In {\em CVPR}, 2022.

\bibitem{ronneberger2015u}
Olaf Ronneberger, Philipp Fischer, and Thomas Brox.
\newblock U-net: Convolutional networks for biomedical image segmentation.
\newblock In {\em MICCAI}, 2015.

\bibitem{russakovsky2015imagenet}
Olga Russakovsky, Jia Deng, Hao Su, Jonathan Krause, Sanjeev Satheesh, Sean Ma,
  Zhiheng Huang, Andrej Karpathy, Aditya Khosla, Michael Bernstein, et~al.
\newblock Imagenet large scale visual recognition challenge.
\newblock {\em IJCV}, 2015.

\bibitem{saharia2022photorealistic}
Chitwan Saharia, William Chan, Saurabh Saxena, Lala Li, Jay Whang, Emily~L
  Denton, Kamyar Ghasemipour, Raphael Gontijo~Lopes, Burcu Karagol~Ayan, Tim
  Salimans, et~al.
\newblock Photorealistic text-to-image diffusion models with deep language
  understanding.
\newblock {\em NeurIPS}, 2022.

\bibitem{salakhutdinov2009deep}
Ruslan Salakhutdinov and Geoffrey Hinton.
\newblock Deep boltzmann machines.
\newblock In {\em Artificial intelligence and statistics (AISTATS)}, 2009.

\bibitem{schuhmann2022laion}
Christoph Schuhmann, Romain Beaumont, Richard Vencu, Cade Gordon, Ross
  Wightman, Mehdi Cherti, Theo Coombes, Aarush Katta, Clayton Mullis, Mitchell
  Wortsman, et~al.
\newblock Laion-5b: An open large-scale dataset for training next generation
  image-text models.
\newblock In {\em NeurIPS}, 2022.

\bibitem{schuhmann2021laion}
Christoph Schuhmann, Richard Vencu, Romain Beaumont, Robert Kaczmarczyk,
  Clayton Mullis, Aarush Katta, Theo Coombes, Jenia Jitsev, and Aran
  Komatsuzaki.
\newblock Laion-400m: Open dataset of clip-filtered 400 million image-text
  pairs.
\newblock {\em arXiv preprint arXiv:2111.02114}, 2021.

\bibitem{sha2023defake}
Zeyang Sha, Zheng Li, Ning Yu, and Yang Zhang.
\newblock De-fake: Detection and attribution of fake images generated by
  text-to-image generation models, 2023.

\bibitem{invisible-watermark}
ShieldMnt.
\newblock invisible-watermark.
\newblock \url{https://github.com/ShieldMnt/invisible-watermark}.

\bibitem{smolensky1986information}
Paul Smolensky et~al.
\newblock Information processing in dynamical systems: Foundations of harmony
  theory.
\newblock 1986.

\bibitem{sohl2015deep}
Jascha Sohl-Dickstein, Eric Weiss, Niru Maheswaranathan, and Surya Ganguli.
\newblock Deep unsupervised learning using nonequilibrium thermodynamics.
\newblock In {\em ICML}, 2015.

\bibitem{song2020denoising}
Jiaming Song, Chenlin Meng, and Stefano Ermon.
\newblock Denoising diffusion implicit models.
\newblock {\em ICLR}, 2021.

\bibitem{midjourneyprompts}
Succinctly.
\newblock Midjourney prompts.
\newblock \url{https://huggingface.co/datasets/succinctly/midjourney-prompts}.

\bibitem{Sudre_2017}
Carole~H. Sudre, Wenqi Li, Tom Vercauteren, Sebastien Ourselin, and M.~Jorge
  Cardoso.
\newblock Generalised dice overlap as a deep learning loss function for highly
  unbalanced segmentations.
\newblock In {\em Deep Learning in Medical Image Analysis and Multimodal
  Learning for Clinical Decision Support}, pages 240--248. Springer
  International Publishing, 2017.

\bibitem{van2016conditional}
Aaron Van~den Oord, Nal Kalchbrenner, Lasse Espeholt, Oriol Vinyals, Alex
  Graves, et~al.
\newblock Conditional image generation with pixelcnn decoders.
\newblock 2016.

\bibitem{van2016pixel}
A{\"a}ron Van Den~Oord, Nal Kalchbrenner, and Koray Kavukcuoglu.
\newblock Pixel recurrent neural networks.
\newblock In {\em ICML}, 2016.

\bibitem{verdoliva2020media}
Luisa Verdoliva.
\newblock Media forensics and deepfakes: an overview.
\newblock {\em IEEE Journal of Selected Topics in Signal Processing},
  14(5):910--932, 2020.

\bibitem{stablediffusion}
Computer Vision and Learning research group at Ludwig Maximilian University~of
  Munich.
\newblock Stable diffusion.
\newblock \url{https://github.com/CompVis/stable-diffusion}.

\bibitem{wang2019detecting}
Sheng-Yu Wang, Oliver Wang, Andrew Owens, Richard Zhang, and Alexei~A Efros.
\newblock Detecting photoshopped faces by scripting photoshop.
\newblock In {\em ICCV}, 2019.

\bibitem{wang2020cnn}
Sheng-Yu Wang, Oliver Wang, Richard Zhang, Andrew Owens, and Alexei~A Efros.
\newblock Cnn-generated images are surprisingly easy to spot... for now.
\newblock In {\em CVPR}, 2020.

\bibitem{wang2023diffusiondb}
Zijie~J. Wang, Evan Montoya, David Munechika, Haoyang Yang, Benjamin Hoover,
  and Duen~Horng Chau.
\newblock Diffusiondb: A large-scale prompt gallery dataset for text-to-image
  generative models, 2023.

\bibitem{yu2015lsun}
Fisher Yu, Ari Seff, Yinda Zhang, Shuran Song, Thomas Funkhouser, and Jianxiong
  Xiao.
\newblock Lsun: Construction of a large-scale image dataset using deep learning
  with humans in the loop.
\newblock {\em arXiv preprint arXiv:1506.03365}, 2015.

\bibitem{yu2022scaling}
Jiahui Yu, Yuanzhong Xu, Jing~Yu Koh, Thang Luong, Gunjan Baid, Zirui Wang,
  Vijay Vasudevan, Alexander Ku, Yinfei Yang, Burcu~Karagol Ayan, et~al.
\newblock Scaling autoregressive models for content-rich text-to-image
  generation.
\newblock {\em TMLR}, 2022.

\bibitem{yun2019cutmix}
Sangdoo Yun, Dongyoon Han, Seong~Joon Oh, Sanghyuk Chun, Junsuk Choe, and
  Youngjoon Yoo.
\newblock Cutmix: Regularization strategy to train strong classifiers with
  localizable features.
\newblock In {\em ICCV}, 2019.

\bibitem{zhang2018unreasonable}
Richard Zhang, Phillip Isola, Alexei~A Efros, Eli Shechtman, and Oliver Wang.
\newblock The unreasonable effectiveness of deep features as a perceptual
  metric.
\newblock In {\em CVPR}, 2018.

\bibitem{zhou2018learning}
Peng Zhou, Xintong Han, Vlad~I Morariu, and Larry~S Davis.
\newblock Learning rich features for image manipulation detection.
\newblock In {\em CVPR}, 2018.

\end{thebibliography}
}

\end{document}